%% file: main.tex
\documentclass[10pt,twocolumn,letterpaper]{article}

\usepackage{cvpr}
\usepackage{times}
\usepackage{epsfig}
\usepackage{graphicx}
\usepackage{amsmath}
\usepackage{amssymb}
\usepackage{booktabs}
\usepackage[font=small]{caption}
\usepackage[font=small]{subcaption}
\usepackage{adjustbox}
\usepackage{comment}
\usepackage{todonotes}
\usepackage{pifont}
\usepackage{multirow}
\usepackage{cite}
\usepackage{enumitem}
\usepackage{authblk}
\usepackage{mathtools}

\usepackage{tikz}
\usetikzlibrary{decorations.pathreplacing}
\usetikzlibrary{shapes.callouts}
\usetikzlibrary{chains, fit, quotes, automata}
\usetikzlibrary{arrows}
\usetikzlibrary{matrix,positioning}

\definecolor{start}{RGB}{255,255,255}
\definecolor{c1}{RGB}{51,160,44}
\definecolor{c2}{RGB}{55,126,184} %
\definecolor{c3}{RGB}{255, 255, 255}
\definecolor{c4}{RGB}{228,26,28}
\definecolor{c5}{RGB}{255,127,0}

\newcommand{\cmark}{\textcolor{c1}{\ding{51}}}%
\newcommand{\xmark}{\textcolor{c4}{\ding{55}}}%

\newcommand{\arch}{\texttt{ESPNetv2}}

\newcommand\pspace{-3mm}

\usepackage[pagebackref=true,breaklinks=true,letterpaper=true,colorlinks,bookmarks=false]{hyperref}

 \cvprfinalcopy 

\def\cvprPaperID{5206} 

\ifcvprfinal\fi
\begin{document}
\providecommand{\mohammad}[1]{{\protect\color{blue}{\bf [Mohammad: #1]}}}


\title{ESPNetv2: A Light-weight, Power Efficient, and General Purpose \\ Convolutional Neural Network}


\author[$\spadesuit$]{Sachin Mehta}
\author[$\heartsuit$ $\clubsuit$]{Mohammad Rastegari}
\author[$\spadesuit$]{Linda Shapiro}
\author[$\spadesuit$ $\heartsuit$]{Hannaneh Hajishirzi}
\affil[ ]{$^\spadesuit$ University of Washington \quad $^\heartsuit$ Allen Institute for AI (AI2) \quad $^\clubsuit$ XNOR.AI}
\affil[ ]{\it {\{sacmehta, shapiro, hannaneh\}@cs.washington.edu}\quad   mohammadr@allenai.org}

\maketitle

\begin{abstract}
   We introduce a light-weight, power efficient, and general purpose convolutional neural network, \arch, for modeling visual and sequential data. Our network uses group point-wise and depth-wise dilated separable convolutions to learn representations from a large effective receptive field with fewer FLOPs and parameters. The performance of our network is evaluated on four different tasks: (1) object classification, (2) semantic segmentation, (3) object detection, and (4) language modeling. Experiments on these tasks, including image classification on the ImageNet and language modeling on the PenTree bank dataset, demonstrate the superior performance of our method over the state-of-the-art methods. Our network outperforms ESPNet by 4-5\% and has $2-4\times$ fewer FLOPs on the PASCAL VOC and the Cityscapes dataset. Compared to YOLOv2 on the MS-COCO object detection, \arch~delivers 4.4\% higher accuracy with $6\times$ fewer FLOPs. Our experiments show that \arch~is much more power efficient than existing state-of-the-art efficient methods including ShuffleNets and MobileNets. Our code is open-source and available at \url{https://github.com/sacmehta/ESPNetv2}. 
\end{abstract}

\section{Introduction}

The increasing programmability and computational power of GPUs have accelerated the growth of deep convolutional neural networks (CNNs) for modeling visual data \cite{krizhevsky2012imagenet,he2016deep,merity2017regularizing}. CNNs are being used in real-world visual recognition applications such as visual scene understanding \cite{zhao2017pyramid} and bio-medical image analysis \cite{ronneberger2015u}. Many of these real-world applications, such as self-driving cars and robots, run on resource-constrained edge devices and demand online processing of data with low latency. 

Existing CNN-based visual recognition systems require large amounts of computational resources, including memory and power. While they achieve high performance on high-end GPU-based machines (e.g. with NVIDIA TitanX), they are often too expensive for resource constrained edge devices such as cell phones and embedded compute platforms. As an example, ResNet-50 \cite{he2016deep}, one of the most well known CNN architecture for image classification, has 25.56 million parameters (98 MB of memory) and performs 2.8 billion high precision operations to classify an image. These numbers are even higher for deeper CNNs, e.g. ResNet-101. These models quickly overtax the limited resources, including compute capabilities, memory, and battery, available on edge devices. 
Therefore, CNNs for real-world applications running on edge devices should be light-weight and efficient while delivering high accuracy.  

Recent efforts for building light-weight networks can be broadly classified as: (1) \textit{Network compression-based methods} remove redundancies in a pre-trained model in order to be more efficient. These models are usually implemented by different parameter pruning techniques \cite{wen2016learning,li2018constrained}. (2) \textit{Low-bit representation-based methods} represent learned weights using few bits instead of high precision floating points \cite{soudry2014expectation,rastegari2016xnor,hubara2016quantized}. These models usually do not change the structure of the network and the convolutional operations could be implemented using logical gates to enable fast processing on CPUs. (3) \textit{Light-weight CNNs} improve the efficiency of a network by factoring computationally expensive convolution operation \cite{howard2017mobilenets, sandler2018mobilenetv2,zhang2017shufflenet, ma2018shufflenet,huang2017condensenet,mehta2018espnet}. These models are computationally efficient by their design i.e. the underlying model structure learns fewer parameters and has fewer floating point operations (FLOPs).       

In this paper, we introduce a light-weight architecture, \arch, that can be easily deployed on edge devices. The main contributions of our paper are:  (1) A general purpose architecture for modeling both visual and sequential data efficiently. We demonstrate the performance of our network across different tasks, ranging from object classification to language modeling. (2) Our proposed architecture, \arch, extends ESPNet \cite{mehta2018espnet}, a dilated convolution-based segmentation network, with depth-wise separable convolutions; an efficient form of convolutions that are used in state-of-art efficient networks including MobileNets \cite{howard2017mobilenets,sandler2018mobilenetv2} and ShuffleNets \cite{ma2018shufflenet,zhang2017shufflenet}. Depth-wise dilated separable convolutions improves the accuracy of \arch~by 1.4\% in comparison to depth-wise separable convolutions. We note that ESPNetv2 achieves better accuracy (72.1 with 284 MFLOPs) with fewer FLOPs than dilated convolutions in the ESPNet \cite{mehta2018espnet} (69.2 with 426 MFLOPs). (3) Our empirical results show that \arch~delivers similar or better performance with fewer FLOPS on different visual recognition tasks. On the ImageNet classification task \cite{ILSVRC15}, our model outperforms all of the previous efficient model designs in terms of efficiency and accuracy, especially under small computational budgets. For example, our model outperforms MobileNetv2 \cite{sandler2018mobilenetv2} by 2\% at a computational budget of 28 MFLOPs. For semantic segmentation on the PASCAL VOC and the Cityscapes dataset, \arch~outperforms ESPNet \cite{mehta2018espnet} by 4-5\% and has $2-4\times$ fewer FLOPs. For object detection, \arch~outperforms YOLOv2 by 4.4\% and has $6\times$ fewer FLOPs. We also study a cyclic learning rate scheduler with warm restarts. Our results suggests that this scheduler is more effective than the standard fixed learning rate scheduler.

\section{Related Work}

\paragraph{Efficient CNN architectures:} Most state-of-the-art efficient networks \cite{howard2017mobilenets, sandler2018mobilenetv2, ma2018shufflenet} use depth-wise separable convolutions \cite{howard2017mobilenets} that factor a convolution into two steps to reduce computational complexity: (1) depth-wise convolution that performs light-weight filtering by applying a single convolutional kernel per input channel and (2) point-wise convolution that usually expands the feature map along channels by learning linear combinations of the input channels. 
Another efficient form of convolution that has been used in efficient networks \cite{zhang2017shufflenet,huang2017condensenet} is group convolution \cite{krizhevsky2012imagenet}, wherein input channels and convolutional kernels are factored into groups and each group is convolved independently. The \arch~network extends the ESPNet network \cite{mehta2018espnet} using these efficient forms of convolutions. To learn representations from a large effective receptive field, \arch~uses depth-wise ``dilated" separable convolutions instead of depth-wise separable convolutions. 

In addition to convolutional factorization, a network's efficiency and accuracy can be further improved using methods such as channel shuffle \cite{ma2018shufflenet} and channel split \cite{ma2018shufflenet}. Such methods are orthogonal to our work.

\vspace{\pspace}
\paragraph{Neural architecture search:} These approaches search over a huge network space using a pre-defined dictionary containing different parameters, including different convolutional layers, different convolutional units, and different filter sizes \cite{zoph2016neural,tan2018mnasnet,cai2018proxylessnas,wu2018fbnet}. Recent search-based methods \cite{tan2018mnasnet,wu2018fbnet} have shown improvements for MobileNetv2. We believe that these methods will increase the performance of \arch~and are complementary to our work.

\vspace{\pspace}
\paragraph{Network compression:} These approaches improve the inference of a pre-trained network by pruning network connections or channels \cite{han2015deep,han2015learning,wen2016learning,li2018constrained,veit2018convolutional}. These approaches are effective, because CNNs have a substantial number of redundant weights. The efficiency gain in most of these approaches are due to the sparsity of parameters, and are difficult to efficiently implement on CPUs due to the cost of look-up and data migration operations. These approaches are complementary to our network.

\vspace{\pspace}
\paragraph{Low-bit representation:} Another approach to improve inference of a pre-trained network is low-bit representation of network weights using quantization  \cite{soudry2014expectation,rastegari2016xnor,wu2016quantized,courbariaux2016binarized,zhou2016dorefa,hubara2016quantized,andri2018yodann}. These approaches use fewer bits to represent weights of a pre-trained network instead of 32-bit high-precision floating points. Similar to network compression-based methods, these approaches are complementary to our work.

\section{\arch}
\label{sec:espnetv2}
This section elaborates the \arch~architecture in detail. We first describe \textit{depth-wise dilated separable convolutions} that enables our network to learn representations from a large effective receptive field efficiently. We then describe the core unit of the \arch~network, the EESP unit, which is built using group point-wise convolutions and depth-wise dilated separable convolutions.
\subsection{Depth-wise dilated separable convolution}

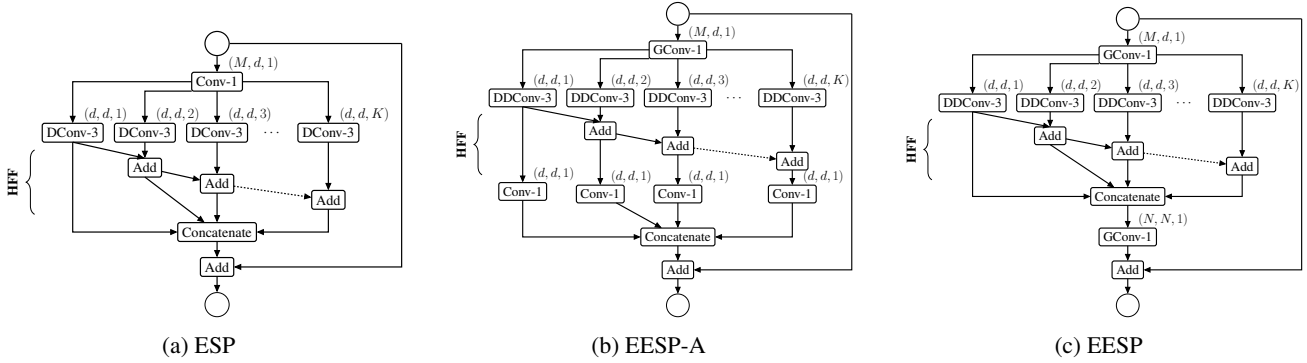
\begin{figure*}[t!]
    \centering
    \begin{subfigure}[b]{0.66\columnwidth}
    \centering
   \resizebox{\columnwidth}{!}{
        \input{tikz/eesp.tikz}\esp
    }
    \caption{ESP}
    \label{fig:esp}
    \end{subfigure}
    \hfill
    \begin{subfigure}[b]{0.66\columnwidth}
    \centering
   \resizebox{\columnwidth}{!}{
        \input{tikz/eesp.tikz}\esppA
    }
    \caption{EESP-A}
    \label{fig:eespa}
    \end{subfigure}
    \hfill
    \begin{subfigure}[b]{0.66\columnwidth}
    \centering
   \resizebox{\columnwidth}{!}{
        \input{tikz/eesp.tikz}\espp
    }
    \caption{EESP}
    \label{fig:eespb}
    \end{subfigure}
    \caption{This figure visualizes the building blocks of the ESPNet, the ESP unit in (a), and the \arch, the EESP unit in (b-c). We note that EESP units in (b-c) are equivalent in terms of computational complexity. Each convolutional layer (Conv-n: $n\times n$ standard convolution, GConv-n: $n\times n$ group convolution, DConv-n: $n\times n$ dilated convolution, DDConv-n: $n\times n$ depth-wise dilated convolution) is denoted by (\# input channels, \# output channels, and dilation rate). Point-wise convolutions in (b) or group point-wise convolutions in (c) are applied after HFF to learn linear combinations between inputs.}
    \label{fig:eespUnit}
\end{figure*}

Convolution factorization is the key principle that has been used by many efficient architectures \cite{howard2017mobilenets, ma2018shufflenet, zhang2017shufflenet, sandler2018mobilenetv2}. The basic idea is to replace the full convolutional operation with a factorized version such as depth-wise separable convolution \cite{howard2017mobilenets} or group convolution \cite{krizhevsky2012imagenet}. In this section, we describe depth-wise dilated separable convolutions and compare with other similar efficient forms of convolution.

A standard convolution convolves an input $\mathbf{X} \in \mathbb{R}^{W \times H \times c}$ with convolutional kernel $\mathbf{K} \in \mathbb{R}^{n \times n \times c \times \hat{c}}$ to produce an output $\mathbf{Y} \in \mathbb{R}^{W \times H \times \hat{c}}$ by learning $n^2 c \hat{c}$ parameters from an effective receptive field of $n \times n$. In contrast to standard convolution, depth-wise dilated separable convolutions apply a light-weight filtering by factoring a standard convolution into two layers: 1) depth-wise dilated convolution per input channel with a dilation rate of $r$; enabling the convolution to learn representations from an effective receptive field of $n_r \times n_r, \text{where } n_r = (n-1)\cdot r + 1$ and 2) point-wise convolution to learn linear combinations of input. This factorization reduces the computational cost by a factor of $\frac{n^2c\hat{c}}{n^2c + c\hat{c}}$. 
A comparison between different types of convolutions is provided in Table \ref{tab:convCompare}. Depth-wise dilated separable convolutions are efficient and can learn representations from large effective receptive fields.

\begin{table}[t!]
    \centering
    \resizebox{\columnwidth}{!}{
    \begin{tabular}{l|c|c}
        \toprule
        \textbf{Convolution type} & \textbf{Parameters} & \textbf{Eff. receptive field}\\
        \midrule        
        Standard & $n^2c\hat{c}$ & $n \times n$\\  
        Group & $\frac{n^2c\hat{c}}{g}$ & $n \times n$\\
        Depth-wise separable & $n^2c + c \hat{c}$ & $n \times n$\\
        Depth-wise dilated separable & $n^2c + c \hat{c}$ & $n_r \times n_r$\\
        \bottomrule
    \end{tabular}
    }
    \caption{Comparison between different type of convolutions. Here, $n\times n$ is the kernel size, $n_r = (n-1)\cdot r + 1$, $r$ is the dilation rate, $c$ and $\hat{c}$ are the input and output channels respectively, and $g$ is the number of groups.}
    \label{tab:convCompare}
\end{table}

\begin{table*}[t!]
    \centering
    \resizebox{1.8\columnwidth}{!}{
    \begin{tabular}{l|c|c|c|c|c|c|c|c|c}
    \toprule
        \multirow{2}{*}{\textbf{Layer}} &  \textbf{Output} & \textbf{Kernel size}  & \multirow{2}{*}{\textbf{Repeat}} & \multicolumn{6}{c}{\multirow{2}{*}{\bf Output channels for different \arch~models}}\\
         &  \textbf{Size} & \textbf{/ Stride}  &  &  \multicolumn{6}{c}{ } \\
    \midrule
    Convolution & 112 $\times$ 112 & 3 $\times$ 3 / 2  & 1 & 16   & 32   & 32   & 32   & 32   & 32   \\
    \midrule
    Strided EESP (Fig. \ref{fig:eespd}) & 56 $\times$ 56 &   & 1 & 32   & 64   & 80   & 96   & 112  & 128  \\
    \midrule
    Strided EESP (Fig. \ref{fig:eespd}) & 28 $\times$ 28 &   & 1 & 64   & 128  & 160  & 192  & 224  & 256  \\
    EESP (Fig. \ref{fig:eespb}) & 28 $\times$ 28 &   & 3 & 64   & 128  & 160  & 192  & 224  & 256  \\
    \midrule
    Strided EESP (Fig. \ref{fig:eespd}) & 14 $\times$ 14 & & 1 & 128  & 256  & 320  & 384  & 448  & 512  \\
    EESP (Fig. \ref{fig:eespb})  & 14 $\times$ 14 &   & 7 & 128  & 256  & 320  & 384  & 448  & 512  \\
    \midrule
    Strided EESP (Fig. \ref{fig:eespd}) & 7 $\times$ 7 &   & 1 & 256  & 512  & 640  & 768  & 896  & 1024 \\
    EESP (Fig. \ref{fig:eespb})  & 7 $\times$ 7 &   & 3 & 256  & 512  & 640  & 768  & 896  & 1024 \\
    Depth-wise convolution & 7 $\times$ 7 & 3 $\times$ 3   &  & 256  & 512  & 640  & 768  & 896  & 1024 \\
    Group convolution & 7 $\times$ 7 & 1 $\times$ 1   &  & 1024 & 1024 & 1024 & 1024 & 1280 & 1280 \\
    \midrule
    Global avg. pool & 1 $\times$ 1 &  7 $\times$ 7& \multicolumn{7}{c}{} \\
    \midrule
    Fully connected &  \multicolumn{3}{c|}{ } & 1000 & 1000 & 1000 & 1000 & 1000 & 1000 \\
    \midrule
    \midrule
    \textbf{Complexity} &  \multicolumn{3}{c|}{ } & 28 M   & 86  M  & 123 M  & 169 M  & 224 M  & 284 M  \\
    \midrule
    \textbf{Parameters} & \multicolumn{3}{c|}{ } & 1.24 M & 1.67 M & 1.97 M & 2.31 M & 3.03 M & 3.49 M \\
    \bottomrule 
    \end{tabular}
    }
    \caption{The \arch~network at different computational complexities for classifying a $224\times 224$ input into 1000 classes in the ImageNet dataset \cite{ILSVRC15}. Network's complexity is evaluated in terms of total number of multiplication-addition operations (or FLOPs).}
    \label{tab:esonetv2}
\end{table*}

\subsection{EESP unit}

Taking advantage of depth-wise dilated separable and group point-wise convolutions, we introduce a new unit EESP, \underline{E}xtremely \underline{E}fficient \underline{S}patial \underline{P}yramid of Depth-wise Dilated Separable Convolutions, which is specifically designed for edge devices. The design of our network is motivated by the ESPNet architecture \cite{mehta2018espnet}, a state-of-the-art efficient segmentation network. The basic building block of the ESPNet architecture is the ESP module, shown in Figure \ref{fig:esp}. It is based on a \textit{reduce-split-transform-merge} strategy. The ESP unit first projects the high-dimensional input feature maps into low-dimensional space using point-wise convolutions and then learn the representations in parallel using dilated convolutions with different dilation rates. Different dilation rates in each branch allow the ESP unit to learn the representations from a large effective receptive field. This factorization, especially learning the representations in a low-dimensional space, allows the ESP unit to be efficient. 

To make the ESP module even more computationally efficient, we first replace point-wise convolutions with group point-wise convolutions. We then replace computationally expensive $3 \times 3$ dilated convolutions with their economical counterparts i.e. depth-wise dilated separable convolutions.  To remove the gridding artifacts caused by dilated convolutions, we fuse the feature maps using the computationally efficient hierarchical feature fusion (HFF) method \cite{mehta2018espnet}. This method additively fuses the feature maps learned using dilated convolutions in a hierarchical fashion; feature maps from the branch with lowest receptive field are combined with the feature maps from the branch with next highest receptive field at each level of the hierarchy\footnote{Other existing works \cite{yu2017dilated,wang2018understanding} add more convolutional layers  with small dilation rates to remove gridding artifacts. This increases the computational complexity of the unit or network.}. The resultant unit is shown in Figure \ref{fig:eespa}. With group point-wise and depth-wise dilated separable convolutions, the total complexity of the ESP block is reduced by a factor of $\frac{Md + n^2d^2K}{\frac{Md}{g} + (n^2 + d)dK}$, where $K$ is the number of parallel branches and $g$ is the number of groups in group point-wise convolution. For example, the EESP unit learns $7\times$ fewer parameters than the ESP unit when $M$=$240$, $g$=$K$=$4$, and $d$=$\frac{M}{K}$=$60$.

We note that computing $K$ point-wise (or $1\times 1$) convolutions in Figure \ref{fig:eespa} independently is equivalent to a single group point-wise convolution with $K$ groups in terms of complexity; however, group point-wise convolution is more efficient in terms of implementation, because it launches one convolutional kernel rather than $K$ point-wise convolutional kernels. Therefore, we replace these $K$ point-wise convolutions with a group point-wise convolution, as shown in Figure \ref{fig:eespb}. We will refer to this unit as EESP.

\begin{figure}[b!]
    \centering
    \centering
   \resizebox{0.75\columnwidth}{!}{
        \input{tikz/eesp.tikz}\esppd
    }
    
    \caption{Strided EESP unit with shortcut connection to an input image (highlighted in red) for down-sampling. The average pooling operation is repeated $P\times$ to match the spatial dimensions of an input image and feature maps.}
    \label{fig:eespd}
\end{figure}
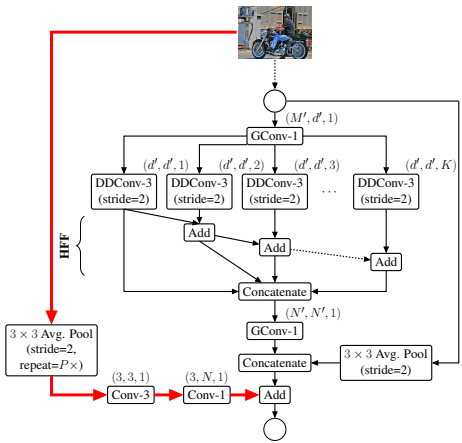

\vspace{\pspace}
\paragraph{Strided EESP with shortcut connection to an input image:} To learn representations efficiently at multiple scales, we make following changes to the EESP block in Figure \ref{fig:eespb}: 1) depth-wise dilated convolutions are replaced with their strided counterpart, 2) an average pooling operation is added instead of an identity connection, and 3) the element-wise addition operation is replaced with a concatenation operation, which helps in expanding the dimensions of feature maps efficiently \cite{zhang2017shufflenet}. 

Spatial information is lost during down-sampling and convolution (filtering) operations. To better encode spatial relationships and learn representations efficiently, we add an efficient long-range shortcut connection between the input image and the current down-sampling unit. This connection first down-samples the image to the same size as that of the feature map and then learns the representations using a stack of two convolutions. The first convolution is a standard $3\times 3$ convolution that learns the spatial representations while the second convolution is a point-wise convolution that learns linear combinations between the input, and projects it to a high-dimensional space. The resultant EESP unit with long-range shortcut connection to the input is shown in Figure \ref{fig:eespd}.

\begin{figure*}[t!]
    \centering
     \begin{subfigure}[b]{0.65\columnwidth}
        \raisebox{-.5\height}{\includegraphics[width=\columnwidth]{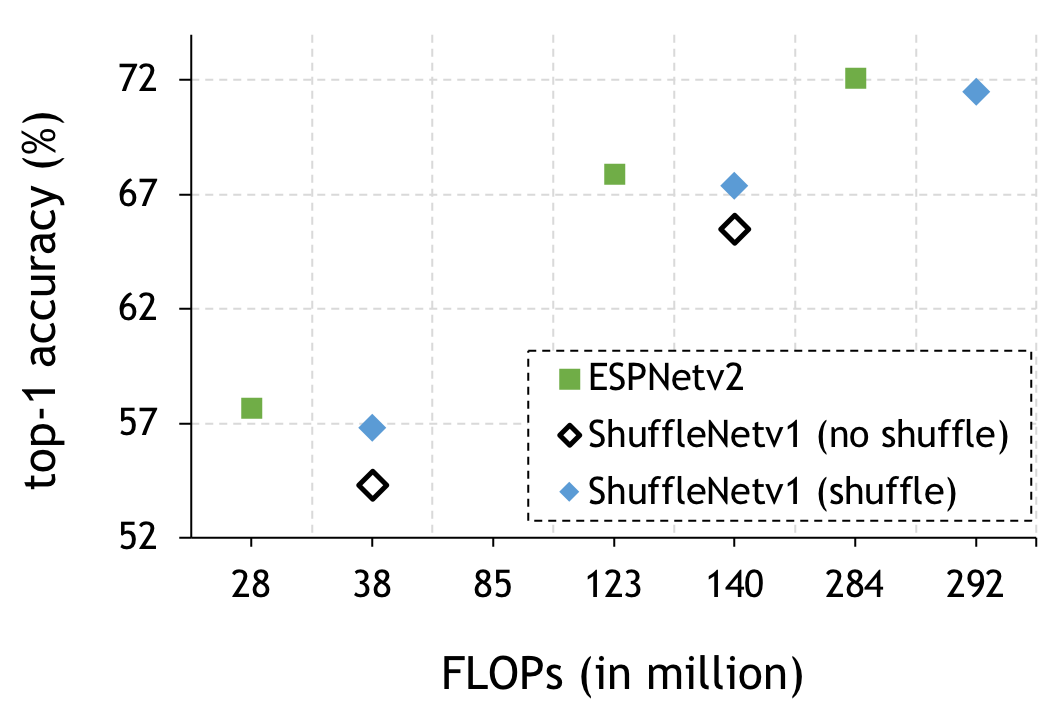}}
    \caption{ }
    \label{fig:shuffCompare}
    \end{subfigure}
    \hfill
    \begin{subfigure}[b]{0.65\columnwidth}
     \raisebox{-.5\height}{\includegraphics[width=\columnwidth]{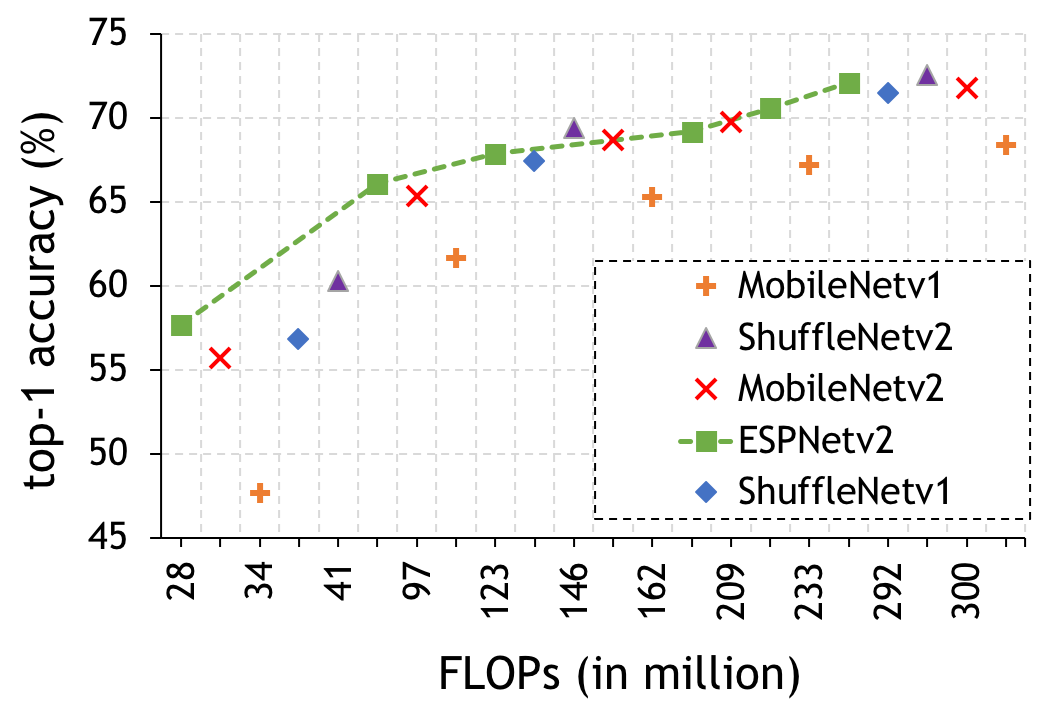}}
    \caption{ }
    \label{fig:shuffCompareA}
    \end{subfigure}
    \hfill
    \begin{subfigure}[b]{0.6\columnwidth}
    \resizebox{\columnwidth}{!}{
        \begin{tabular}{l|r|r|r}
        \toprule
            \textbf{Network} & \textbf{\# Params} & \textbf{FLOPs} & \textbf{Top-1}  \\
        \midrule        
            MobileNetv1 \cite{howard2017mobilenets} & 2.59 M & 325 M  & 68.4 \\
            CondenseNet \cite{huang2017condensenet} & -- & 274 M  & 71.0 \\
            IGCV3 \cite{sun2018igcv3} & --& 318 M  & 72.2 \\
            Xception$^\dagger$ \cite{chollet2017xception} & -- & 305 M & 70.6\\
            DenseNet$^\dagger$ \cite{huang2017densely} & -- & 295 M & 60.1 \\
            ShuffleNetv1 \cite{zhang2017shufflenet} & 3.46 M & 292 M & 71.5\\
            \midrule
            \multirow{2}{*}{MobileNetv2 \cite{sandler2018mobilenetv2}} & 3.47 M  & 300 M  & 71.8 \\
            & 6.9 M  & 585 M  & 74.7 \\
            \midrule
            \multirow{2}{*}{ ShuffleNetv2 \cite{ma2018shufflenet}} & 3.5 M & 299 M & 72.6\\
            & 7.4 M & 591 M & 74.9 \\
            \midrule
            \multirow{2}{*}{\arch~(Ours)} & 3.49 M & 284 M & 72.1 \\
             & 5.9 M & 602 M & 74.9 \\
            \bottomrule
        \end{tabular}
    }
    \caption{ }
    \end{subfigure}
    \caption{Performance comparison of different efficient networks on the ImageNet validation set: (a) \arch~vs. ShuffleNetv1 \cite{zhang2017shufflenet}, (b) \arch~vs. efficient models at different network complexities, and (c) \arch~vs. state-of-the-art for a computational budget of approximately 300 million FLOPs. We count the total number of multiplication-addition operations (FLOPs) for an input image of size $224\times 224$. Here, $^\dagger$ represents that the performance of these networks is reported in \cite{ma2018shufflenet}. Best viewed in color.}
    \label{fig:effCompare}
\end{figure*}

\subsection{Network architecture}
The \arch~network is built using EESP units. At each spatial level, the \arch~repeats the EESP units several times to increase the depth of the network. In the EESP unit (Figure \ref{fig:eespb}), we use batch normalization \cite{ioffe2015batch} and PReLU \cite{he2015delving} after every convolutional layer with an exception to the last group-wise convolutional layer where PReLU is applied after element-wise sum operation. To maintain the same computational complexity at each spatial-level, the feature maps are doubled after every down-sampling operation \cite{simonyan2014very, he2016deep}. 

In our experiments, we set the dilation rate $r$ proportional to the number of branches in the EESP unit ($K$). The effective receptive field of the EESP unit grows with $K$. Some of the kernels, especially at low spatial levels such as $7\times7$, might have a larger effective receptive field than the size of the feature map. Therefore, such kernels might not contribute to learning. In order to have meaningful kernels, we limit the effective receptive field at each spatial level $l$ with spatial dimension $W^l \times H^l$ as: $\text{n}_d^l(Z^l) = 5 + \frac{Z^l}{7}, Z^l \in \{W^l, H^l\}$ with the effective receptive field ($n_d \times n_d$) corresponding to the lowest spatial level (i.e. $7\times7$) as $5 \times 5$. Following \cite{mehta2018espnet}, we set $K=4$ in our experiments. Furthermore, in order to have a homogeneous architecture, we set the number of groups in group point-wise convolutions equal to number of parallel branches ($g=K$). The overall \arch~architectures at different computational complexities are shown in Table \ref{tab:esonetv2}.

\section{Experiments}
To showcase the power of the \arch~network, we evaluate and compare the performance with state-of-the-art methods on four different tasks: (1) object classification, (2) semantic segmentation, (3) object detection, and (3) language modeling.
\subsection{Image classification}
\label{ssec:imagenet}

\paragraph{Dataset:} We evaluate the performance of the \arch~on the ImageNet 1000-way classification dataset \cite{ILSVRC15} that contains 1.28M images for training and 50K images for validation. We evaluate the performance of our network using the single crop top-1 classification accuracy, i.e. we compute the accuracy on the center cropped view of size $224 \times 224$.

\begin{figure}[t!]
    \centering
    \includegraphics[width=0.65\columnwidth]{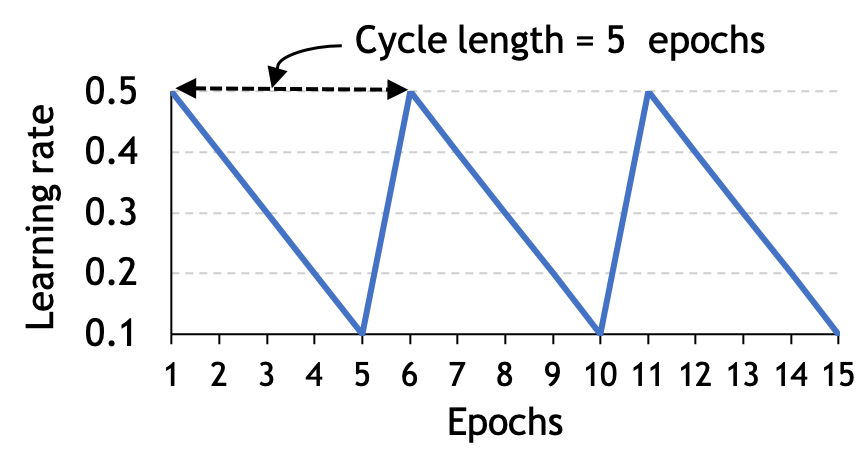}
    \caption{Cyclic learning rate policy (see Eq.\ref{eq:clr}) with linear learning rate decay and warm restarts.}
    \label{fig:clr}
\end{figure}

\vspace{\pspace}
\paragraph{Training:} The \arch~networks are trained using the PyTorch deep learning framework \cite{paszke2017automatic} with CUDA 9.0 and cuDNN as the back-ends. For optimization, we use SGD \cite{sutskever2013importance} with \textit{warm restarts}. At each epoch $t$, we compute the learning rate $\eta_t$ as:
\begin{equation}
    \eta_t = \eta_{max} - (t\ \text{mod}\ T) \cdot \eta_{min}
\label{eq:clr}
\end{equation}
where $\eta_{max}$ and $\eta_{min}$ are the ranges for the learning rate and $T$ is the cycle length after which learning rate will restart. Figure \ref{fig:clr} visualizes the learning rate policy for three cycles. This learning rate scheme can be seen as a variant of the cosine learning policy \cite{loshchilov2016sgdr}, wherein the learning rate is decayed as a function of cosine before warm restart. In our experiment, we set $\eta_{min}=0.1$, $\eta_{max}=0.5$, and $T=5$. We train our networks with a batch size of 512 for 300 epochs by optimizing the cross-entropy loss. For faster convergence, we  decay the learning rate by a factor of two at the following epoch intervals:  \{50, 100, 130, 160, 190, 220, 250, 280\}. We use a standard data augmentation strategy \cite{szegedy2015going,he2016deep} with an exception to color-based normalization. This is in contrast to recent efficient architectures that uses less scale augmentation to prevent under-fitting \cite{ma2018shufflenet, zhang2017shufflenet}. The weights of our networks are initialized using the method described in \cite{he2015delving}.

\begin{figure*}[t!]
    \centering
    \begin{subfigure}[b]{0.69\columnwidth}
        \includegraphics[width=\columnwidth]{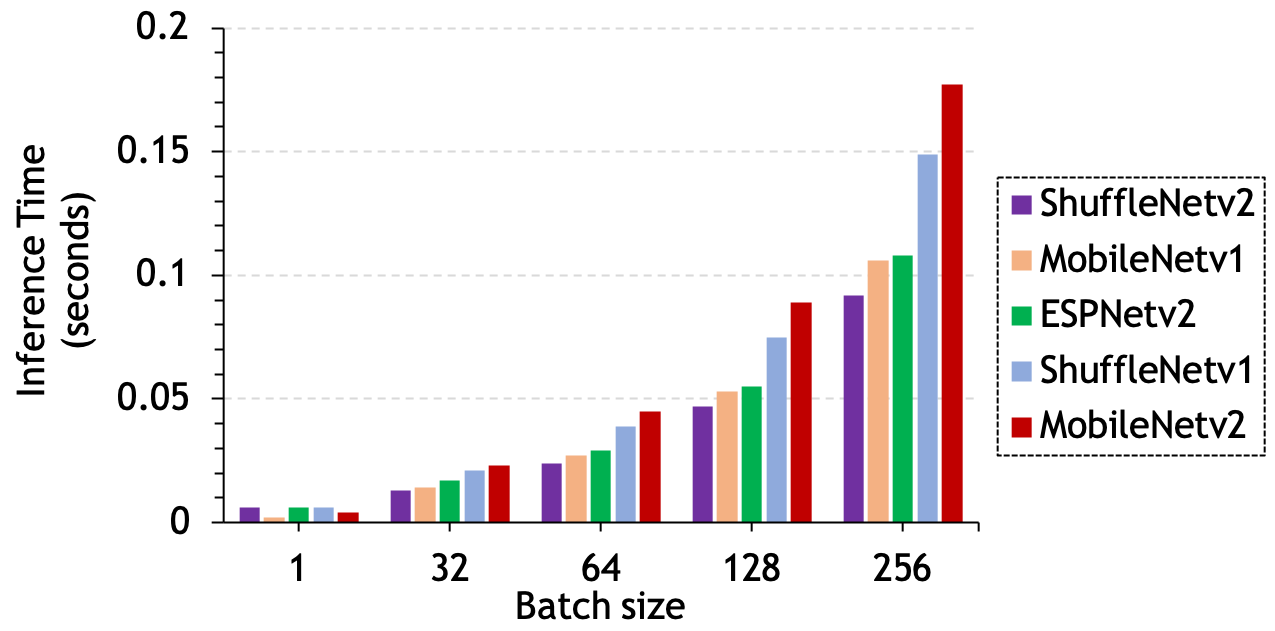}
        \caption{Inference time vs. batch size (1080 Ti)}
    \end{subfigure}
    \hfill
    \begin{subfigure}[b]{0.69\columnwidth}
        \includegraphics[width=\columnwidth]{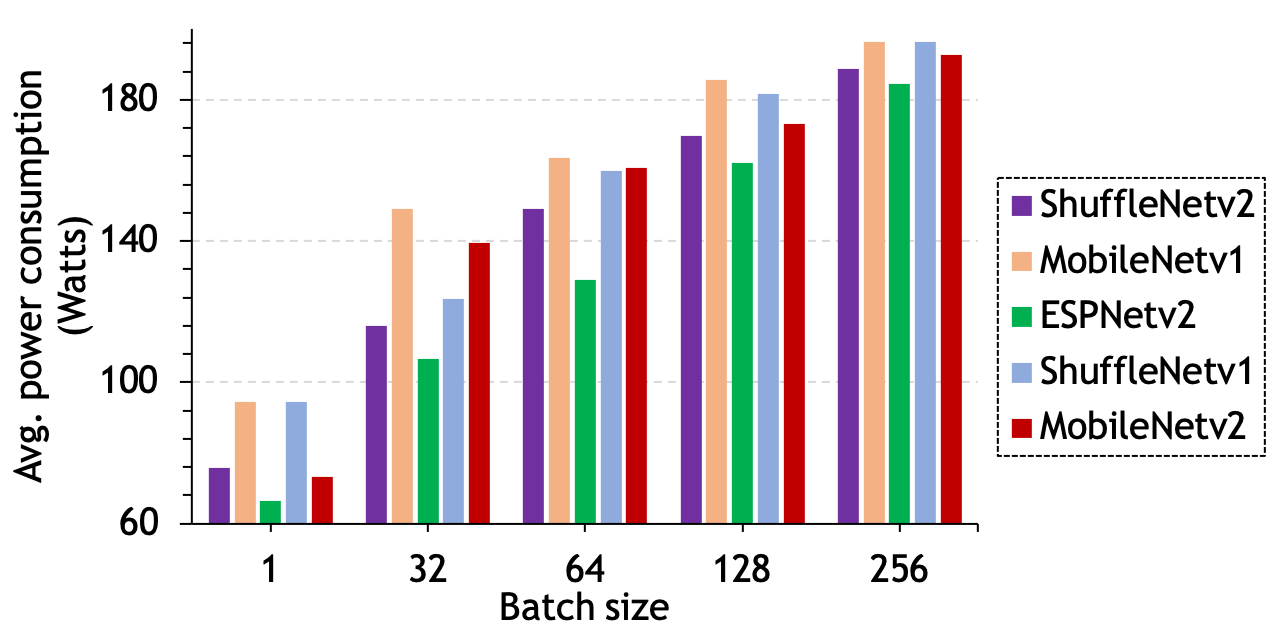}
        \caption{Power vs. batch size (1080 Ti)}
    \end{subfigure}
    \hfill
    \begin{subfigure}[b]{0.62\columnwidth}
        \includegraphics[width=\columnwidth]{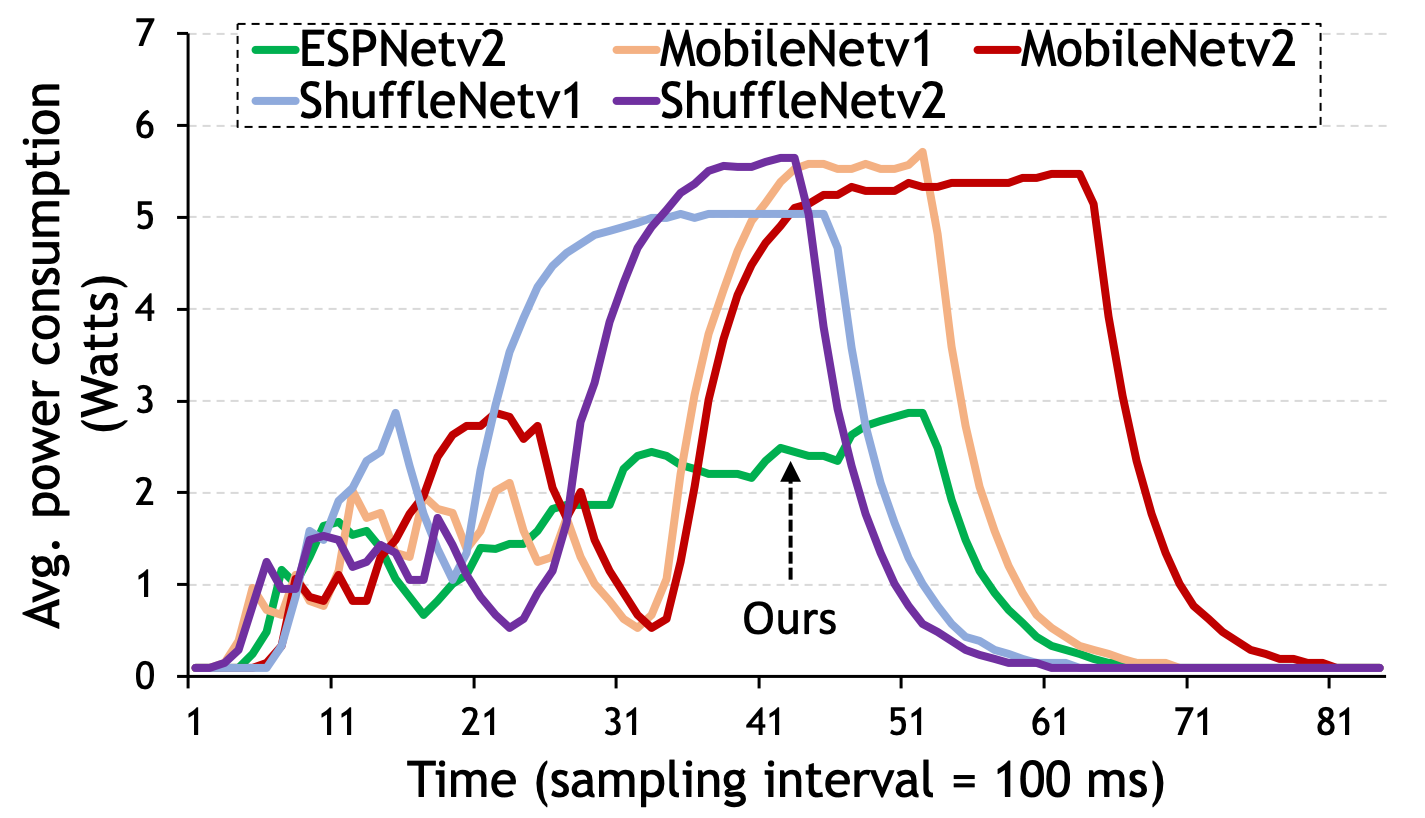}
        \caption{Power consumption on TX2}
    \end{subfigure}
    \caption{Performance analysis of different efficient networks (computational budget $\approx$ 300 million FLOPs). Inference time and power consumption are averaged over 100 iterations for a $224\times 224$ input on a NVIDIA GTX 1080 Ti GPU and NVIDIA Jetson TX2. We do not report execution time on TX2 because there is not much substantial difference. Best viewed in color.}
    \label{fig:perfAnalysis}
\end{figure*}

\vspace{\pspace}
\paragraph{Results:} Figure \ref{fig:effCompare} provides a performance comparison between \arch~and state-of-the-art efficient networks. We observe that: (1) Like ShuffleNetv1 \cite{zhang2017shufflenet}, \arch~also uses group point-wise convolutions. However, \arch~does not use any channel shuffle which was found to be very effective in ShuffleNetv1 and delivers better performance than ShuffleNetv1. (2) Compared to MobileNets, \arch~delivers better performance especially under small computational budgets. With 28 million FLOPs, \arch~outperforms MobileNetv1 \cite{howard2017mobilenets}  (34 million FLOPs) and MobileNetv2 \cite{sandler2018mobilenetv2} (30 million FLOPs) by 10\% and 2\% respectively. (3) \arch~delivers comparable accuracy to ShuffleNetv2 \cite{ma2018shufflenet} without any channel split, which enables ShuffleNetv2 to deliver better performance than ShuffleNetv1. We believe that such functionalities (channel split and channel shuffle) are orthogonal to \arch~and can be used to further improve its efficiency and accuracy. (4) Compared to other efficient networks at a computational budget of about 300 million FLOPs, \arch~delivered better performance (e.g. 1.1\% more accurate than the CondenseNet \cite{huang2017condensenet}). 

\begin{figure}[t!]
    \centering
    \includegraphics[width=0.65\columnwidth]{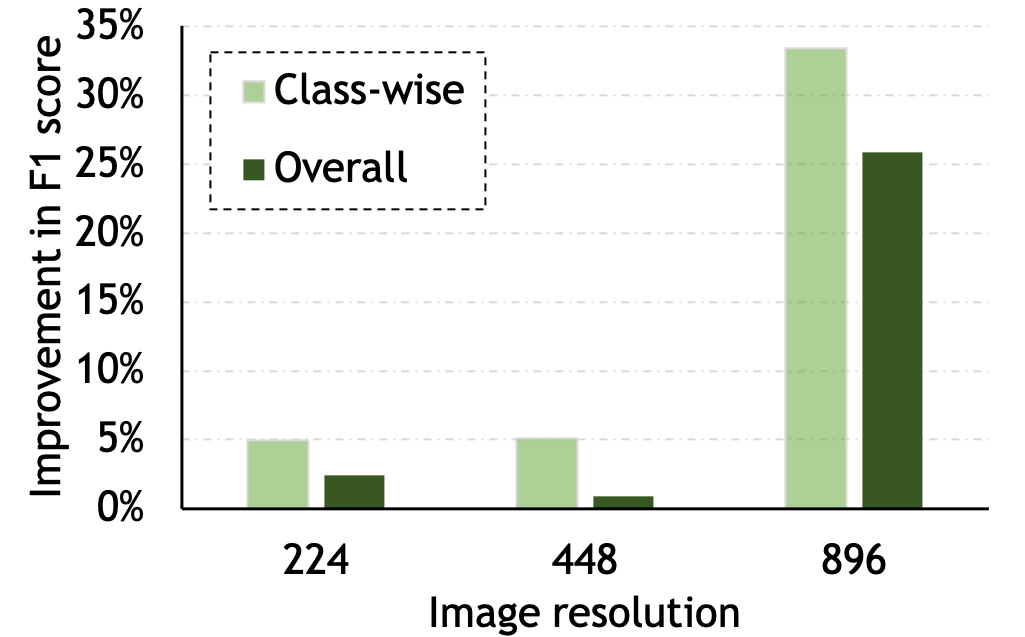}
    \caption{Performance improvement in F1-score of \arch~over ShuffleNetv2 on MS-COCO multi-object classification task when \textit{tested} at different image resolutions. Class-wise/overall F1-scores for \arch~and ShuffleNetv2 for an input of $224\times224$ on the validation set are 63.41/69.23 and 60.42/67.58 respectively.}
    \label{fig:coco}
\end{figure}

\vspace{\pspace}
\paragraph{Multi-label classification:} To evaluate the generalizability for transfer learning, we evaluate our model on the MSCOCO multi-object classification task \cite{lin2014microsoft}. The dataset consists of 82,783 images, which are categorized into 80 classes with 2.9 object labels per image. Following \cite{zhu2017learning}, we evaluated our method on the validation set (40,504 images) using class-wise and overall F1 score. We finetune \arch~(284 million FLOPs) and Shufflenetv2 \cite{ma2018shufflenet} (299 million FLOPs) for 100 epochs using the same data augmentation and training settings as for the ImageNet dataset, except $\eta_{max}$=$0.005$, $\eta_{min}$=$0.001$ and learning rate is decayed by two at the 50th and 80th epochs. We use binary cross entropy loss for optimization. Results are shown in Figure \ref{fig:coco}. \arch~outperforms ShuffleNetv2 by a large margin, especially when tested at image resolution of $896\times 896$; suggesting large effective receptive fields of the EESP unit help \arch~learn better representations.

\vspace{\pspace}
\paragraph{Performance analysis:} Edge devices have limited computational resources and restrictive energy overhead. An efficient network for such devices should consume less power and have low latency with a high accuracy. We measure the efficiency of our network, \arch, along with other state-of-the-art networks (MobileNets \cite{howard2017mobilenets, sandler2018mobilenetv2} and ShuffleNets \cite{zhang2017shufflenet, ma2018shufflenet}) on two different devices: 1) a high-end graphics card (NVIDIA GTX 1080 Ti) and 2) an embedded device (NVIDIA Jetson TX2). For a fair comparison, we use PyTorch as a deep-learning framework. Figure \ref{fig:perfAnalysis} compares the inference time and power consumption while networks complexity along with their accuracy are compared in Figure \ref{fig:effCompare}. The inference speed of \arch~is slightly lower than the fastest network (ShuffleNetv2 \cite{ma2018shufflenet}) on both devices, however, it is much more power efficient while delivering similar accuracy on the ImageNet dataset. This suggests that \arch~network has a good trade-off between accuracy, power consumption, and latency; a much desirable property for any network running on edge devices.

\subsection{Semantic segmentation}
\label{ssec:semseg}

\paragraph{Dataset:} We evaluate the performance of the \arch~on two datasets: (1) the Cityscapes \cite{cordts2016cityscapes} and (2) the PASCAL VOC 2012 dataset \cite{pascal2012}. The Cityscapes dataset consists of 5,000 finely annotated images (training/validation/test: 2,975/500/1,525). The task is to segment an image into 19 classes that belongs to 7 categories. The PASCAL VOC 2012 dataset provide annotations for 20 foreground objects and has 1.4K training, 1.4K validation, and 1.4K test images. Following a standard convention \cite{chen2016deeplab,zhao2017pspnet}, we also use additional images from \cite{hariharan2011semantic,lin2014microsoft} for training our networks.

\vspace{\pspace}
\paragraph{Training:} We train our network in two stages. In the first stage, we use a smaller image resolution for training ($256\times256$ for the PASCAL VOC 2012 dataset and $512\times256$ for the CityScapes dataset). We train \arch~for 100 epochs using SGD with an initial learning rate of 0.007. In the second stage, we increase the image resolution ($384\times384$ for the PASCAL VOC 2012 and $1024\times512$ for the Cityscapes dataset) and then finetune the \arch~from first stage for 100 epochs using SGD with initial learning rate of 0.003. For both these stages, we use cyclic learning schedule discussed in Section \ref{ssec:imagenet}. For the first 50 epochs, we use a cycle length of 5 while for the remaining epochs, we use a cycle length of 50 i.e. for the last 50 epochs, we decay the learning rate linearly. We evaluate the accuracy in terms of mean Intersection over Union (mIOU) on the private test set using \textit{online evaluation server}. For evaluation, we up-sample segmented masks to the same size as of the input image using nearest neighbour interpolation.

\begin{figure}[b!]
    \centering
    \begin{subfigure}[b]{0.51\columnwidth}
        \centering
        \resizebox{\columnwidth}{!}{
            \begin{tabular}{lrr}
                \toprule
                \textbf{Network} & \multicolumn{1}{l}{\textbf{FLOPs}} & \multicolumn{1}{l}{\textbf{mIOU}} \\
                \midrule
                SegNet \cite{badrinarayanan2017segnet} & 82 B & 57.0 \\
                ContextNet \cite{poudel2018contextnet} & 33 B & 66.1 \\
                ICNet \cite{zhao2018icnet} & 31 B & 69.5 \\
                ERFNet \cite{romera2018erfnet} & 26 B & 69.7 \\
                MobileNetv2$^{\star\star}$ \cite{sandler2018mobilenetv2}  & 21 B & \textbf{70.7}\\
                \midrule
                RTSeg- MobileNet \cite{siam2018rtseg} & 13.8 B & 61.5 \\
                RTSeg-ShuffleNet \cite{siam2018rtseg} & 6.2 B & 58.3 \\
                ESPNet \cite{mehta2018espnet} & 4.5 B & 60.3 \\
                ENet \cite{paszke2016enet} & 3.8 B & 58.3 \\
                \midrule
                \arch-val (Ours) & \textbf{2.7 B} & 66.4\\ 
                \arch-test (Ours) & \textbf{2.7 B} & 66.2\\ 
                \bottomrule
            \end{tabular}
        }
        \caption{Cityscapes}
        \label{fig:compareSt}
    \end{subfigure}
    \hfill
    \begin{subfigure}[b]{0.47\columnwidth}
        \centering
        \resizebox{\columnwidth}{!}{
        \begin{tabular}{lrr}
            \toprule
            \textbf{Network} & \textbf{FLOPs} & \textbf{mIOU} \\
            \midrule
            FCN-8s \cite{long2015fully} & 181 B & 62.2 \\
            DeepLabv3 \cite{chen2017rethinking} & 81 B & \textbf{80.49}\\
            SegNet \cite{badrinarayanan2017segnet} & 31 B & 59.1 \\
            \midrule
            MobileNetv1 \cite{howard2017mobilenets} & 14 B & 75.29 \\
            MobileNetv2 \cite{sandler2018mobilenetv2} & 5.8 B & 75.7 \\
            ESPNet \cite{mehta2018espnet} & 2.2 B & 63.01 \\
            \midrule
            \arch~- val & 0.76 B & 67.0 \\
             \arch~- test & \textbf{0.76 B} & 68.0 \\
            \bottomrule
        \end{tabular}
        }
        \caption{PASCAL VOC 2012}
    \end{subfigure}
\caption{Semantic segmentation results on (a) the Cityscapes dataset and (b) the PASCAL VOC 2012 dataset. For a fair comparison, we report FLOPs at the same image resolution which is used for computing the accuracy. \\ {\scriptsize 
$^{\star\star}$ \cite{sandler2018mobilenetv2} uses additional data from \cite{lin2014microsoft}
}}
\label{fig:compareCityScapes}
\end{figure}

\vspace{\pspace}
\paragraph{Results:} Figure \ref{fig:compareCityScapes} compares the performance of \arch~with state-of-the-methods on both the Cityscapes and the PASCAL VOC 2012 dataset. We can see that \arch~delivers a competitive performance to existing methods while being very efficient. Under the similar computational constraints, \arch~outerperforms existing methods like ENet and ESPNet by large margin. Notably, \arch~is 2-3\% less accurate than other efficient networks such as ICNet, ERFNet, and ContextNet, but has $9-12\times$ fewer FLOPs.

\subsection{Object detection}
\paragraph{Dataset and training details:} For object detection, we replace VGG with \arch~in single shot object detector. We evaluate the performance on two datasets: (1) the PASCAL VOC 2007 and (2) the MS-COCO dataset. For the PASCAL VOC 2007 dataset, we also use additional images from the PASCAL VOC 2012 dataset. We evaluate the performance in terms of mean Average Precision (mAP). For the COCO dataset, we report mAP @ IoU of 0.50:0.95. For training, we use the same learning policy as in Section \ref{ssec:semseg}.

\vspace{\pspace}
\paragraph{Results:} Table \ref{table:objDet} compares the performance of \arch~with existing methods. \arch~provides a good trade-off between accuracy and efficiency. Notably, \arch~delivers the same performance as YOLOv2, but has $25\times$ fewer FLOPs. Compared to SSD, \arch~delivers a very competitive performance while being very efficient.

\begin{table}[t!]
    \centering
        \resizebox{0.8\columnwidth}{!}{
        \begin{tabular}{l|r|r|r|r}
            \toprule
            \multirow{2}{*}{\textbf{Network}} & \multicolumn{2}{c|}{\textbf{VOC07}} & \multicolumn{2}{c}{\textbf{COCO}} \\
             & \textbf{FLOPs} & \textbf{mAP} & \textbf{FLOPs} & \textbf{mAP} \\
            \midrule
            SSD-512 \cite{liu2016ssd}  & 90.2 B & \textbf{74.9} & 99.5 B & \textbf{26.8} \\
            SSD-300 \cite{liu2016ssd} & 31.3 B & 72.4 & 35.2 B & 23.2 \\
            YOLOv2 \cite{redmon2017yolo9000} & 6.8 B & 69.0 & 17.5 B & 21.6 \\
            \midrule
            MobileNetv1-320 \cite{howard2017mobilenets} & -- & -- & 1.3 B & 22.2 \\
            MobileNetv2-320 \cite{sandler2018mobilenetv2} & -- & -- & 0.8 B & 22.1 \\
            \midrule
            \arch-512 (Ours) & 2.5 B & 68.2 & 2.8 B & 26.0 \\
            \arch-384 (Ours) & 1.4 B & 65.6 & 1.6 B & 23.2 \\
            \arch-256 (Ours) & \textbf{0.6 B} & 63.8 & \textbf{0.7 B} & 21.9 \\
            \bottomrule
        \end{tabular}
        }
    \caption{Object detection results on the PASCAL VOC 2007 and the MS-COCO dataset.}
    \label{table:objDet}
\end{table}

\subsection{Language modeling}

\paragraph{Dataset:} The performance of our unit, the EESP, is evaluated on the Penn Treebank (PTB) dataset \cite{marcus1993building} as prepared by \cite{mikolov2010recurrent}. For training and evaluation, we follow the same splits of training, validation, and test data as in \cite{merity2017regularizing}.

\vspace{\pspace}
\paragraph{Language Model:} We extend LSTM-based language models by replacing linear transforms for processing the input vector with the EESP unit inside the LSTM cell\footnote{We replace 2D convolutions with 1D convolutions in the EESP unit.}. We call this model ERU (Efficient Recurrent Unit). Our model uses 3-layers of ERU with an embedding size of 400. We use standard dropout \cite{srivastava2014dropout} with probability of 0.5 after embedding layer, the output between ERU layers, and the output of final ERU layer. We train the network using the same learning policy as \cite{merity2017regularizing}. We evaluate the performance in terms of perplexity; a lower value of perplexity is desirable. 

\vspace{\pspace}
\paragraph{Results:} Language modeling results are provided in Table \ref{tab:lmERU}. ERUs achieve similar or better performance than state-of-the-art methods while learning fewer parameters. With similar hyper-parameter settings such as dropout, ERUs deliver similar (only 1 point less than PRU \cite{mehta2018espnet}) or better performance than state-of-the-art recurrent networks while learning fewer parameters; suggesting that the introduced EESP unit (Figure \ref{fig:eespb}) is efficient and powerful, and can be applied across different sequence modeling tasks such as question answering and machine translation. We note that our smallest language model with 7 million parameters outperforms most of state-of-the-art language models (e.g. \cite{gal2016theoretically,Chen2018alter,bradbury2016quasi}). We believe that the performance of ERU can be further improved by rigorous hyper-parameter search \cite{melis2017state} and advanced dropouts \cite{merity2017regularizing,gal2016theoretically}.

\begin{table}[t!]
    \centering
    \resizebox{0.9\columnwidth}{!}{
    \begin{tabular}{l|c|c}
    \toprule
        \textbf{Language Model} & \# \textbf{Params} & \textbf{Perplexity} \\
    \midrule
        Variational LSTM \cite{gal2016theoretically} & 20 M & 78.6 \\
        SRU \cite{lei2017sru} & 24 M &  60.3\\
        Quantized LSTM \cite{Chen2018alter} & -- & 89.8 \\
        QRNN \cite{bradbury2016quasi} & 18 M &  78.3 \\
        Skip-connection LSTM \cite{melis2017state} & 24 M &  58.3 \\
        AWD-LSTM \cite{merity2017regularizing} & 24 M & 57.3 \\
        PRU \cite{mehta2018pyramidal} (with standard dropout \cite{srivastava2014dropout}) & 19 M & 62.42 \\
        AWD-PRU \cite{mehta2018pyramidal} (with weight dropout \cite{merity2017regularizing})  & 19 M & \textbf{56.56} \\
        \midrule
        \multirow{2}{*}{ERU-Ours (with standard dropout \cite{srivastava2014dropout})}& \textbf{7 M} & 73.63 \\
         & 15 M & 63.47\\
        \bottomrule
    \end{tabular}
    }
    \caption{This table compares single model word-level perplexity of our model with state-of-the-art on test set of the Penn Treebank dataset. Lower perplexity value represents better performance.}
    \label{tab:lmERU}
\end{table}

\section{Ablation Studies on the ImageNet Dataset}
\label{sec:ablations}
This section elaborate on various choices that helped make \arch~efficient and accurate.

\vspace{\pspace}
\paragraph{Impact of different convolutions:} Table \ref{tab:diffconvs} summarizes the impact of different convolutions. Clearly, depth-wise dilated separable convolutions are more effective than dilated and depth-wise convolutions.

\begin{table}[b!]
    \centering
    \resizebox{0.7\columnwidth}{!}{
    \begin{tabular}{l|r|r}
        \toprule
        \textbf{Convolution} & \textbf{FLOPs} & \textbf{top-1}  \\
        \midrule
        Dilated (standard)  &  478 M &  69.2 \\
        Depth-wise separable & 123 M & 66.5\\
        Depth-wise dilated separable  & 123 M & 67.9 \\
        \bottomrule
    \end{tabular}
    }
    \caption{\arch~with different convolutions. ESPNetv2 with standard dilated convolutions is the same as ESPNet.}
    \label{tab:diffconvs}
\end{table}

\vspace{\pspace}
\paragraph{Impact of hierarchical feature fusion (HFF):} In \cite{mehta2018espnet}, HFF is introduced  to remove gridding artifacts caused by dilated convolutions. Here, we study their influence on object classification. The performance of the \arch~network with and without HFF are shown in Table \ref{tab:ablation} (see R1 and R2). HFF improves classification performance by about 1.5\% while having no impact on the network's complexity. This suggests that the role of HFF is dual purpose. First, it removes gridding artifacts caused by dilated convolutions (as noted by \cite{mehta2018espnet}). Second, it enables sharing of information between different branches of the EESP unit (see Figure \ref{fig:eespb}) that allows it to learn rich and strong representations.

\vspace{\pspace}
\paragraph{Impact of long-range shortcut connections with the input:} To see the influence of shortcut connections with the input image, we train the \arch~network with and without shortcut connection. Results are shown in Table \ref{tab:ablation} (see R2 and R3). Clearly, these connections are effective and efficient, improving the performance by about 1\% with a little (or negligible) impact on network's complexity.

\begin{table}[t!]
    \centering
    \resizebox{\columnwidth}{!}{
    \begin{tabular}{c|c|c|c|c|c|c|c}
    \toprule
        & \multicolumn{2}{|c|}{\bf Network properties} & \multicolumn{2}{|c|}{\bf Learning schedule} & \multicolumn{3}{c}{\bf Performance}\\
    \cline{2-8}
        & \textbf{HFF} & \textbf{LRSC} & \textbf{Fixed} & \textbf{Cyclic} & \textbf{\# Params} & \textbf{FLOPs} & \textbf{Top-1} \\
    \cline{2-8}
        R1 & \xmark & \xmark & \cmark & \xmark & 1.66 M & 84 M & 58.94 \\
        R2 & \cmark & \xmark & \cmark & \xmark & 1.66 M & 84 M & 60.07 \\
        R3 & \cmark & \cmark & \cmark & \xmark & 1.67 M & 86 M & 61.20 \\
        R4 & \cmark & \cmark & \xmark & \cmark & 1.67 M & 86 M & 62.17 \\
        R5$^\dagger$ & \cmark & \cmark & \xmark & \cmark & 1.67 M & 86 M & 66.10 \\
    \bottomrule
    \end{tabular}
    }
    \caption{Performance of \arch~under different settings. Here, HFF represents hierarchical feature fusion and LRSC represents long-range shortcut connection with an input image. We train \arch~for 90 epochs and decay the learning rate by 10 after every 30 epochs. For fixed learning rate schedule, we initialize learning rate with 0.1 while for cyclic, we set $\eta_{min}$ and $\eta_{max}$ to 0.1 and 0.5 in Eq. \ref{eq:clr} respectively. Here, $^\dagger$ represents that the learning rate schedule is the same as in Section \ref{ssec:imagenet}.}
    \label{tab:ablation}
\end{table}

\vspace{\pspace}
\paragraph{Fixed vs cyclic learning schedule:} A comparison between fixed and cyclic learning schedule is shown in Table \ref{tab:ablation} (R3 and R4). With cyclic learning schedule, the \arch~network achieves about 1\% higher top-1 validation accuracy on the ImageNet dataset; suggesting that cyclic learning schedule allows to find a better local minima than fixed learning schedule. Further, when we trained \arch~network for longer (300 epochs) using the learning schedule outlined in Section \ref{ssec:imagenet}, performance improved by about 4\% (see R4 and R5 in Table \ref{tab:ablation}). 

\section{Conclusion}
We introduce a light-weight and power efficient network, \arch, which better encode the spatial information in images by learning representations from a large effective receptive field. Our network is a general purpose network with good generalization abilities and can be used across a wide range of tasks, including sequence modeling. Our network delivered state-of-the-art performance across different tasks such as object classification, detection, segmentation, and language modeling while being more power efficient. 

\vspace{2mm}
\begin{small}
\noindent {\bf Acknowledgement:} This research was supported by the Intelligence Advanced Research Projects Activity (IARPA) via Interior/Interior Business Center (DOI/IBC) contract number D17PC00343, NSF III (1703166), Allen Distinguished Investigator Award, Samsung GRO award, and gifts from Google, Amazon, and Bloomberg. We also thank Rik Koncel-Kedziorski, David Wadden, Beibin Li, and Anat Caspi for their helpful comments. The U.S. Government is authorized to reproduce and distribute reprints for Governmental purposes notwithstanding any copyright annotation thereon. Disclaimer: The views and conclusions contained herein are those of the authors and should not be interpreted as necessarily representing endorsements, either expressed or implied, of IARPA, DOI/IBC, or the U.S. Government.
\end{small}


{\small
\bibliographystyle{ieee}
\bibliography{main}
}

\end{document}

%% file: tikz/eesp.tikz
\tikzset{>=triangle 45}
\newcommand\lw{0.75mm}

\newcommand{\esp}{
  \begin{tikzpicture}[
block/.style={
inner sep=8,outer sep=0,
align=center,rounded corners, line width=\lw,
font=\Huge}]

	\node[circle, draw, minimum size=1.5cm, line width=\lw, fill=start] (x) {\quad};
    \node[circle, fill=white, right=10.5cm of x] (dummy1) {\quad};
    \node[block, draw, fill=white, below=1cm of x.south, label=above right:{\Huge$(M, d, 1)$}] (xDot) {$\text{Conv-1}$};

    \node[block,draw, fill=white, below=2cm of xDot, label=above right:{\Huge$(d, d, 3)$}] (x11) {$\text{DConv-3}$};
    \node[block, right=0.7cm of x11] (xDot1) {\Huge $\cdots$};
    \node[block,draw, fill=white, left=0.7cm of x11, label=above right:{\Huge$(d, d, 2)$}] (x21) {$\text{DConv-3}$};
    \node[block,draw, fill=white, left=0.7cm of x21, label=above right:{\Huge$(d, d, 1)$}] (x41) {$\text{DConv-3}$};
    \node[block,draw, fill=white,  right=0.7cm of xDot1, label=above right:{\Huge$(d, d, K)$}] (x31) {$\text{DConv-3}$};
    
    \node[fill=white, below left=0.25cm of x41,align=center,line width=\lw] (dumx41a) {};
    \node[fill=white, below=4cm of dumx41a,align=center,line width=\lw] (dumx41) {};
    \draw [decorate,decoration={brace,amplitude=15pt,mirror,raise=4pt},yshift=2pt]
(dumx41a) -- (dumx41) node [black,midway,xshift=-1.5cm, rotate=90] {\Huge \textbf{HFF}};
    
    \node[block,draw, fill=white, below=1cm of x21] (dum0) {\Huge Add};
    \node[block,draw, fill=white, below=2cm of x11] (dum1) {\Huge Add};
    \node[block,draw, fill=white, below=3cm of x31] (dum2) {\Huge Add};

    \draw[thick, ->, line width=\lw] (x41.south) -- (dum0.north);
    \draw[thick, ->, line width=\lw] (x21.south) -- (dum0);
    \draw[thick, ->, line width=\lw] (x11.south) -- (dum1.north);
    \draw[thick, ->, line width=\lw] (dum0) -- (dum1);
    \draw[dashed, thick, ->, line width=\lw] (dum1) -- (dum2);
    \draw[thick, ->, line width=\lw] (x31.south) -- (dum2.north);
 
    \node[block,draw, fill=white, below=5cm of x11] (sum1) {\Huge Concatenate};

    \node[block,draw, fill=white, below=1cm of sum1] (sum2) {\Huge Add};
    
    \node[circle, draw, minimum size=1.5cm, fill=start, below=1cm of sum2] (y) {\quad};
    
    \draw[thick, ->, line width=\lw] (x.south) -- (xDot);
    
    \draw[thick, ->, line width=\lw] (xDot.south) -- (x11);
    \draw[thick, ->, line width=\lw] (xDot.south west) -| (x21.north);
    \draw[thick, ->, line width=\lw] (xDot.east) -| (x31.north);
    \draw[thick, ->, line width=\lw] (xDot.west) -| (x41.north);
    
    \draw[thick, ->, line width=\lw] (x41.south) |- (sum1);
    \draw[thick, ->, line width=\lw] (dum0.south) -- (sum1);
    \draw[thick, ->, line width=\lw] (dum1.south) -- (sum1);
    \draw[thick, ->, line width=\lw] (dum2.south) |- (sum1);
    
    \draw[thick, ->, line width=\lw] (sum1) -- (sum2);
    
    \draw[thick, line width=\lw] (x.east) -- (dummy1.center);
    \draw[thick, ->, line width=\lw] (dummy1.center) |- (sum2);
    \draw[thick, ->, line width=\lw] (sum2) -- (y.north);
 \end{tikzpicture}
}

\newcommand{\espAAAAA}{
  \begin{tikzpicture}[
block/.style={
inner sep=8,outer sep=0,
align=center,rounded corners, line width=\lw,
font=\Huge}]

	\node[circle, draw, minimum size=1.5cm, line width=\lw, fill=start] (x) {\quad};
    \node[circle, fill=white, right=10.5cm of x] (dummy1) {\quad};
    \node[block, draw, fill=white, below=1cm of x.south, label=above right:{\Huge$(M, d, 1)$}] (xDot) {$\text{Conv-1}$};

    \node[block,draw, fill=white, below=2cm of xDot, label=above right:{\Huge$(d, d, 3)$}] (x11) {$\text{DConv-3}$};
    \node[block, right=0.7cm of x11] (xDot1) {\Huge $\cdots$};
    \node[block,draw, fill=white, left=0.7cm of x11, label=above right:{\Huge$(d, d, 2)$}] (x21) {$\text{DConv-3}$};
    \node[block,draw, fill=white, left=0.7cm of x21, label=above right:{\Huge$(d, d, 1)$}] (x41) {$\text{DConv-3}$};
    \node[block,draw, fill=white,  right=0.7cm of xDot1, label=above right:{\Huge$(d, d, K)$}] (x31) {$\text{DConv-3}$};
    
    \node[fill=white, below left=0.25cm of x41,align=center,line width=\lw] (dumx41a) {};
    \node[fill=white, below=4cm of dumx41a,align=center,line width=\lw] (dumx41) {};
    \draw [decorate,decoration={brace,amplitude=15pt,mirror,raise=4pt},yshift=2pt]
(dumx41a) -- (dumx41) node [black,midway,xshift=-1.5cm, rotate=90] {\Huge \textbf{HFF}};
    
    \node[block,draw, fill=white, below=1cm of x21] (dum0) {\Huge Add};
    \node[block,draw, fill=white, below=2cm of x11] (dum1) {\Huge Add};
    \node[block,draw, fill=white, below=3cm of x31] (dum2) {\Huge Add};

    \draw[thick, ->, line width=\lw] (x41.south) -- (dum0.north);
    \draw[thick, ->, line width=\lw] (x21.south) -- (dum0);
    \draw[thick, ->, line width=\lw] (x11.south) -- (dum1.north);
    \draw[thick, ->, line width=\lw] (dum0) -- (dum1);
    \draw[dashed, thick, ->, line width=\lw] (dum1) -- (dum2);
    \draw[thick, ->, line width=\lw] (x31.south) -- (dum2.north);
 
    \node[block,draw, fill=white, below=5cm of x11] (sum1) {\Huge Concatenate};

    \node[block,draw, fill=white, below=1cm of sum1] (sum2) {\Huge Add};
    
    \node[circle, draw, minimum size=1.5cm, fill=start, below=1cm of sum2] (y) {\quad};
    
    \draw[thick, ->, line width=\lw] (x.south) -- (xDot);
    
    \draw[thick, ->, line width=\lw] (xDot.south) -- (x11);
    \draw[thick, ->, line width=\lw] (xDot.south west) -| (x21.north);
    \draw[thick, ->, line width=\lw] (xDot.east) -| (x31.north);
    \draw[thick, ->, line width=\lw] (xDot.west) -| (x41.north);
    
    \draw[thick, ->, line width=\lw] (x41.south) |- (sum1);
    \draw[thick, ->, line width=\lw] (dum0.south) -- (sum1);
    \draw[thick, ->, line width=\lw] (dum1.south) -- (sum1);
    \draw[thick, ->, line width=\lw] (dum2.south) |- (sum1);
    
    \draw[thick, ->, line width=\lw] (sum1) -- (sum2);
    
    \draw[thick, line width=\lw] (x.east) -- (dummy1.center);
    \draw[thick, ->, line width=\lw] (dummy1.center) |- (sum2);
    \draw[thick, ->, line width=\lw] (sum2) -- (y.north);
 \end{tikzpicture}
}

\newcommand{\esppA}{
  \begin{tikzpicture}[
block/.style={
inner sep=8,outer sep=0,
align=center,rounded corners, line width=\lw,
font=\Huge}]

	\node[circle, draw, minimum size=1.5cm, line width=\lw, fill=start] (x) {\quad};
    \node[circle, fill=white, right=10.5cm of x] (dummy1) {\quad};
    \node[block, draw, fill=white, below=1cm of x.south, label=above right:{\Huge$(M, d, 1)$}] (xDot) {$\text{GConv-1}$};

    \node[block,draw, fill=white, below=2cm of xDot, label=above right:{\Huge$(d, d, 3)$}] (x11) {$\text{DDConv-3}$};
    \node[block, right=0.7cm of x11] (xDot1) {\Huge $\cdots$};
    \node[block,draw, fill=white, left=0.7cm of x11, label=above right:{\Huge$(d, d, 2)$}] (x21) {$\text{DDConv-3}$};
    \node[block,draw, fill=white, left=0.7cm of x21, label=above right:{\Huge$(d, d, 1)$}] (x41) {$\text{DDConv-3}$};
    \node[block,draw, fill=white,  right=0.7cm of xDot1, label=above right:{\Huge$(d, d, K)$}] (x31) {$\text{DDConv-3}$};
    
    \node[fill=white, below left=0.25cm of x41,align=center,line width=\lw] (dumx41a) {};
    \node[fill=white, below=4cm of dumx41a,align=center,line width=\lw] (dumx41) {};
    \draw [decorate,decoration={brace,amplitude=15pt,mirror,raise=4pt},yshift=2pt]
(dumx41a) -- (dumx41) node [black,midway,xshift=-1.5cm, rotate=90] {\Huge \textbf{HFF}};
    
    \node[block,draw, fill=white, below=1cm of x21] (dum0) {\Huge Add};
    \node[block,draw, fill=white, below=2cm of x11] (dum1) {\Huge Add};
    \node[block,draw, fill=white, below=3cm of x31] (dum2) {\Huge Add};

    \node[block,draw, fill=white, below=5cm of x41, label=above right:{\Huge$(d, d, 1)$}] (conv0) {\Huge Conv-1};
    \node[block,draw, fill=white, below=3cm of dum0, label=above right:{\Huge$(d, d, 1)$}] (conv1) {\Huge Conv-1};
    \node[block,draw, fill=white, below=2cm of dum1, label=above right:{\Huge$(d, d, 1)$}] (conv2) {\Huge Conv-1};
    \node[block,draw, fill=white, below=1cm of dum2, label=above right:{\Huge$(d, d, 1)$}] (conv3) {\Huge Conv-1};

    \draw[thick, ->, line width=\lw] (x41.south) -- (dum0.north);
    \draw[thick, ->, line width=\lw] (x21.south) -- (dum0);
    \draw[thick, ->, line width=\lw] (x11.south) -- (dum1.north);
    \draw[thick, ->, line width=\lw] (dum0) -- (dum1);
    \draw[dashed, thick, ->, line width=\lw] (dum1) -- (dum2);
    \draw[thick, ->, line width=\lw] (x31.south) -- (dum2.north);
 
    \node[block,draw, fill=c3, below=8cm of x11] (sum1) {\Huge Concatenate};

    \node[block,draw, fill=c3, below=1cm of sum1] (sum2) {\Huge Add};
    
    \node[circle, draw, minimum size=1.5cm, fill=start, below=1cm of sum2] (y) {\quad};
    
    \draw[thick, ->, line width=\lw] (x.south) -- (xDot);
    
    \draw[thick, ->, line width=\lw] (xDot.south) -- (x11);
    \draw[thick, ->, line width=\lw] (xDot.south west) -| (x21.north);
    \draw[thick, ->, line width=\lw] (xDot.east) -| (x31.north);
    \draw[thick, ->, line width=\lw] (xDot.west) -| (x41.north);
    
    \draw[thick, ->, line width=\lw] (x41.south) -| (conv0);
    \draw[thick, ->, line width=\lw] (dum0.south) -- (conv1);
    \draw[thick, ->, line width=\lw] (dum1.south) -- (conv2);
    \draw[thick, ->, line width=\lw] (dum2.south) -| (conv3);

    \draw[thick, ->, line width=\lw] (conv0) |- (sum1);
    \draw[thick, ->, line width=\lw] (conv1)-- (sum1);
    \draw[thick, ->, line width=\lw] (conv2)-- (sum1);
    \draw[thick, ->, line width=\lw] (conv3) |- (sum1);
    
    \draw[thick, ->, line width=\lw] (sum1) -- (sum2);
    
    \draw[thick, line width=\lw] (x.east) -- (dummy1.center);
    \draw[thick, ->, line width=\lw] (dummy1.center) |- (sum2);
    \draw[thick, ->, line width=\lw] (sum2) -- (y.north);
 \end{tikzpicture}
}

\newcommand{\espp}{
  \begin{tikzpicture}[
block/.style={
inner sep=8,outer sep=0,
align=center,rounded corners, line width=\lw,
font=\Huge}]

	\node[circle, draw, minimum size=1.5cm, line width=\lw, fill=start] (x) {\quad};
    \node[circle, fill=white, right=10.5cm of x] (dummy1) {\quad};
    \node[block, draw, fill=white, below=1cm of x.south, label=above right:{\Huge$(M, d, 1)$}] (xDot) {$\text{GConv-1}$};

    \node[block,draw, fill=white, below=2cm of xDot, label=above right:{\Huge$(d, d, 3)$}] (x11) {$\text{DDConv-3}$};
    \node[block, right=0.7cm of x11] (xDot1) {\Huge $\cdots$};
    \node[block,draw, fill=white, left=0.7cm of x11, label=above right:{\Huge$(d, d, 2)$}] (x21) {$\text{DDConv-3}$};
    \node[block,draw, fill=white, left=0.7cm of x21, label=above right:{\Huge$(d, d, 1)$}] (x41) {$\text{DDConv-3}$};
    \node[block,draw, fill=white,  right=0.7cm of xDot1, label=above right:{\Huge$(d, d, K)$}] (x31) {$\text{DDConv-3}$};
    
    \node[fill=white, below left=0.25cm of x41,align=center,line width=\lw] (dumx41a) {};
    \node[fill=white, below=4cm of dumx41a,align=center,line width=\lw] (dumx41) {};
    \draw [decorate,decoration={brace,amplitude=15pt,mirror,raise=4pt},yshift=2pt]
(dumx41a) -- (dumx41) node [black,midway,xshift=-1.5cm, rotate=90] {\Huge \textbf{HFF}};
    
    \node[block,draw, fill=white, below=1cm of x21] (dum0) {\Huge Add};
    \node[block,draw, fill=white, below=2cm of x11] (dum1) {\Huge Add};
    \node[block,draw, fill=white, below=3cm of x31] (dum2) {\Huge Add};
    
    \draw[thick, ->, line width=\lw] (x41.south) -- (dum0.north);
    \draw[thick, ->, line width=\lw] (x21.south) -- (dum0);
    \draw[thick, ->, line width=\lw] (x11.south) -- (dum1.north);
    \draw[thick, ->, line width=\lw] (dum0) -- (dum1);
    \draw[dashed, thick, ->, line width=\lw] (dum1) -- (dum2);
    \draw[thick, ->, line width=\lw] (x31.south) -- (dum2.north);
 
    \node[block,draw, fill=white, below=5cm of x11] (sum1) {\Huge Concatenate};

    \node[block,draw, fill=white, minimum size=0.1cm, below=1.5cm of sum1, line width=\lw, label=above right:{\Huge$(N, N, 1)$}] (gOut) {$\text{GConv-1}$};

    \node[block,draw, fill=white, below=1cm of gOut] (sum2) {\Huge Add};
    
    \node[circle, draw, minimum size=1.5cm, fill=start, below=1cm of sum2] (y) {\quad};
    
    \draw[thick, ->, line width=\lw] (x.south) -- (xDot);
    
    \draw[thick, ->, line width=\lw] (xDot.south) -- (x11);
    \draw[thick, ->, line width=\lw] (xDot.south west) -| (x21.north);
    \draw[thick, ->, line width=\lw] (xDot.east) -| (x31.north);
    \draw[thick, ->, line width=\lw] (xDot.west) -| (x41.north);
    
    \draw[thick, ->, line width=\lw] (x41.south) |- (sum1);
    \draw[thick, ->, line width=\lw] (dum0.south) -- (sum1);
    \draw[thick, ->, line width=\lw] (dum1.south) -- (sum1);
    \draw[thick, ->, line width=\lw] (dum2.south) |- (sum1);
    
    \draw[thick, ->, line width=\lw] (sum1) -- (gOut);
    \draw[thick, ->, line width=\lw] (gOut) -- (sum2);
    
    \draw[thick, line width=\lw] (x.east) -- (dummy1.center);
    \draw[thick, ->, line width=\lw] (dummy1.center) |- (sum2);
    \draw[thick, ->, line width=\lw] (sum2) -- (y.north);
 \end{tikzpicture}
}

\newcommand{\esppd}{
  \begin{tikzpicture}[
block/.style={
inner sep=8,outer sep=0,
align=center,rounded corners, line width=\lw,
font=\Huge}]

	\node[circle, draw, minimum size=1.5cm, line width=\lw, fill=start] (x) {\quad};
\node[circle, fill=white, right=11.5cm of x] (dummy1) {\quad};
    \node[above=2cm of x] (img) {\includegraphics[width=140px, height=100px]{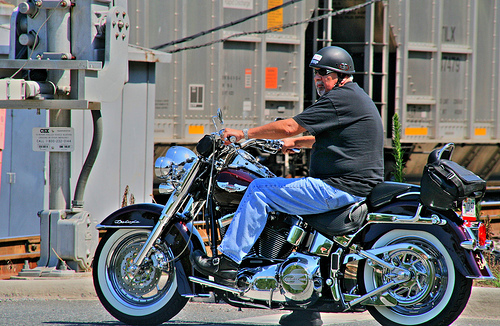}};
    
    \draw[thick, dashed, ->, line width=\lw] (img) -- (x);

    \node[block, draw, fill=white, below=1cm of x.south, label=above right:{\Huge$(M', d', 1)$}] (xDot) {$\text{GConv-1}$};

    \node[block,draw, fill=white, below=2cm of xDot, label=above right:{\Huge$(d', d', 3)$}] (x11) {$\text{DDConv-3}$\\(stride=2)};
    \node[block, right=0.7cm of x11] (xDot1) {\Huge $\cdots$};
    \node[block,draw, fill=white, left=0.7cm of x11, label=above right:{\Huge$(d', d', 2)$}] (x21) {$\text{DDConv-3}$\\(stride=2)};
    \node[block,draw, fill=white, left=0.7cm of x21, label=above right:{\Huge$(d', d', 1)$}] (x41) {$\text{DDConv-3}$\\(stride=2)};
    \node[block,draw, fill=white,  right=0.7cm of xDot1, label=above right:{\Huge$(d', d', K)$}] (x31) {$\text{DDConv-3}$\\(stride=2)};
    
    \node[fill=white, below left=0.25cm of x41,align=center,line width=\lw] (dumx41a) {};
    \node[fill=white, below=4cm of dumx41a,align=center,line width=\lw] (dumx41) {};
    \draw [decorate,decoration={brace,amplitude=15pt,mirror,raise=4pt},yshift=2pt]
(dumx41a) -- (dumx41) node [black,midway,xshift=-1.5cm, rotate=90] {\Huge \textbf{HFF}};
    
    \node[block,draw, fill=c3, below=1cm of x21] (dum0) {\Huge Add};
    \node[block,draw, fill=c3, below=2cm of x11] (dum1) {\Huge Add};
    \node[block,draw, fill=c3, below=3cm of x31] (dum2) {\Huge Add};
    
    \draw[thick, ->, line width=\lw] (x41.south) -- (dum0.north);
    \draw[thick, ->, line width=\lw] (x21.south) -- (dum0);
    \draw[thick, ->, line width=\lw] (x11.south) -- (dum1.north);
    \draw[thick, ->, line width=\lw] (dum0) -- (dum1);
    \draw[dashed, thick, ->, line width=\lw] (dum1) -- (dum2);
    \draw[thick, ->, line width=\lw] (x31.south) -- (dum2.north);
 
    \node[block,draw, fill=white, below=5cm of x11] (sum1) {\Huge Concatenate};

    \node[block,draw, fill=white, minimum size=0.1cm, below=1.5cm of sum1, line width=\lw, label=above right:{\Huge$(N', N', 1)$}] (gOut) {$\text{GConv-1}$};

    \node[block,draw, fill=c3, below=1cm of gOut] (sum2) {\Huge Concatenate};
    \node[block,draw, fill=c3, below=1cm of sum2] (sum3) {\Huge Add};
    \node[block, draw, fill=c3, left=2cm of sum3, label=above:{\Huge$(3, N, 1)$}] (imgB1) {$\text{Conv-1}$};
    \node[block, draw, fill=c3, left=2cm of imgB1, label=above:{\Huge$(3, 3, 1)$}] (imgB2) {$\text{Conv-3}$};
    \node[block, draw, fill=c3, above left=1cm of imgB2] (imgBAvg) {\Huge $3\times3$ Avg. Pool \\(stride=2, \\repeat=$P\times$)};

    \draw[thick, solid, ->, line width=2mm, color=red] (img) -| (imgBAvg);
    \draw[thick, solid, ->, line width=2mm, color=red] (imgBAvg) |- (imgB2);
    \draw[thick, solid, ->, line width=2mm, color=red] (imgB2) -- (imgB1);
\draw[thick, solid, ->, line width=2mm, color=red] (imgB1) |- (sum3);

    \node[block,draw, fill=c3, right=2cm of sum2] (avg) {\Huge $3\times3$ Avg. Pool \\(stride=2)};
    
    \node[circle, draw, minimum size=1.5cm, fill=start, below=1cm of sum3] (y) {\quad};
    
    \draw[thick, ->, line width=\lw] (x.south) -- (xDot);
    
    \draw[thick, ->, line width=\lw] (xDot.south) -- (x11);
    \draw[thick, ->, line width=\lw] (xDot.south west) -| (x21.north);
    \draw[thick, ->, line width=\lw] (xDot.east) -| (x31.north);
    \draw[thick, ->, line width=\lw] (xDot.west) -| (x41.north);
    
    \draw[thick, ->, line width=\lw] (x41.south) |- (sum1);
    \draw[thick, ->, line width=\lw] (dum0.south) -- (sum1);
    \draw[thick, ->, line width=\lw] (dum1.south) -- (sum1);
    \draw[thick, ->, line width=\lw] (dum2.south) |- (sum1);
    
    \draw[thick, ->, line width=\lw] (sum1) -- (gOut);
    \draw[thick, ->, line width=\lw] (gOut) -- (sum2);
    
    \draw[thick, line width=\lw] (x.east) -- (dummy1.center);
    \draw[thick, ->, line width=\lw] (dummy1.center) |- (avg);
    \draw[thick, ->, line width=\lw] (avg) -- (sum2);
    \draw[thick, ->, line width=\lw] (sum2) -- (sum3);
    \draw[thick, ->, line width=\lw] (sum3) -- (y.north);

 \end{tikzpicture}
}

\newcommand{\esppdA}{
  \begin{tikzpicture}[
block/.style={
inner sep=8,outer sep=0,
align=center,rounded corners, line width=\lw,
font=\Huge}]

	\node[circle, draw, minimum size=1.5cm, line width=\lw, fill=start] (x) {\quad};
\node[above=2cm of x] (img) {\includegraphics[width=140px, height=100px]{tikz/example.jpg}};

    \draw[thick, dashed, ->, line width=\lw] (img) -- (x);

    \node[circle, fill=white, right=11cm of x] (dummy1) {\quad};
    \node[block, draw, fill=c1, below=1cm of x.south, label=above right:{\Huge$(M, d, 1)$}] (xDot) {$\text{GConv-1}$};

    \node[block,draw, fill=c2, below=2cm of xDot, label=above right:{\Huge$(d, d, 3)$}] (x11) {$\text{DDConv-3}$\\(stride=2)};
    \node[block, right=0.7cm of x11] (xDot1) {\Huge $\cdots$};
    \node[block,draw, fill=c2, left=0.7cm of x11, label=above right:{\Huge$(d, d, 2)$}] (x21) {$\text{DDConv-3}$\\(stride=2)};
    \node[block,draw, fill=c2, left=0.7cm of x21, label=above right:{\Huge$(d, d, 1)$}] (x41) {$\text{DDConv-3}$\\(stride=2)};
    \node[block,draw, fill=c2,  right=0.7cm of xDot1, label=above right:{\Huge$(d, d, K)$}] (x31) {$\text{DDConv-3}$\\(stride=2)};
    
    \node[fill=white, below left=0.25cm of x41,align=center,line width=\lw] (dumx41a) {};
    \node[fill=white, below=4cm of dumx41a,align=center,line width=\lw] (dumx41) {};
    \draw [decorate,decoration={brace,amplitude=15pt,mirror,raise=4pt},yshift=2pt]
(dumx41a) -- (dumx41) node [black,midway,xshift=-1.5cm, rotate=90] {\Huge \textbf{HFF}};
    
    \node[block,draw, fill=c3, below=1cm of x21] (dum0) {\Huge Add};
    \node[block,draw, fill=c3, below=2cm of x11] (dum1) {\Huge Add};
    \node[block,draw, fill=c3, below=3cm of x31] (dum2) {\Huge Add};
    
    \draw[thick, ->, line width=\lw] (x41.south) -- (dum0.north);
    \draw[thick, ->, line width=\lw] (x21.south) -- (dum0);
    \draw[thick, ->, line width=\lw] (x11.south) -- (dum1.north);
    \draw[thick, ->, line width=\lw] (dum0) -- (dum1);
    \draw[dashed, thick, ->, line width=\lw] (dum1) -- (dum2);
    \draw[thick, ->, line width=\lw] (x31.south) -- (dum2.north);
 
    \node[block,draw, fill=c3, below=5cm of x11] (sum1) {\Huge Concatenate};

    \node[block,draw, fill=c1, minimum size=0.1cm, below=1.5cm of sum1, line width=\lw, label=above right:{\Huge$(N, N, 1)$}] (gOut) {$\text{GConv-1}$};

    \node[block,draw, fill=c3, below=1cm of gOut] (sum2) {\Huge Concatenate};
    \node[block,draw, fill=c3, right=2cm of sum2] (avg) {\Huge $3\times3$ Avg. Pool \\(stride=2)};
    
    \node[circle, draw, minimum size=1.5cm, fill=start, below=1cm of sum2] (y) {\quad};
    
    \draw[thick, ->, line width=\lw] (x.south) -- (xDot);
    
    \draw[thick, ->, line width=\lw] (xDot.south) -- (x11);
    \draw[thick, ->, line width=\lw] (xDot.south west) -| (x21.north);
    \draw[thick, ->, line width=\lw] (xDot.east) -| (x31.north);
    \draw[thick, ->, line width=\lw] (xDot.west) -| (x41.north);
    
    \draw[thick, ->, line width=\lw] (x41.south) |- (sum1);
    \draw[thick, ->, line width=\lw] (dum0.south) -- (sum1);
    \draw[thick, ->, line width=\lw] (dum1.south) -- (sum1);
    \draw[thick, ->, line width=\lw] (dum2.south) |- (sum1);
    
    \draw[thick, ->, line width=\lw] (sum1) -- (gOut);
    \draw[thick, ->, line width=\lw] (gOut) -- (sum2);
    
    \draw[thick, line width=\lw] (x.east) -- (dummy1.center);
    \draw[thick, ->, line width=\lw] (dummy1.center) |- (avg);
    \draw[thick, ->, line width=\lw] (avg) -- (sum2);
    \draw[thick, ->, line width=\lw] (sum2) -- (y.north);

 \end{tikzpicture}
}